%% file: main.tex
\documentclass[a4paper,fleqn]{cas-dc}

\usepackage[numbers,sort&compress]{natbib}
\usepackage{lmodern}
\usepackage{amsmath}
\usepackage{booktabs}
\usepackage{threeparttable}
\usepackage{graphicx}
\usepackage{subcaption}
\usepackage{placeins}
\usepackage{xurl}

\graphicspath{{fig/}}

\AtBeginDocument{\DeclareFontShape{T1}{lmr}{b}{it}{<->ssub*lmr/bx/it}{}}

\hypersetup{
  breaklinks=true,
  colorlinks=true,
  linkcolor=blue,
  citecolor=blue,
  urlcolor=blue
}

\ExplSyntaxOn
\RenewDocumentCommand \printorcid { }
{
  \seq_if_empty:NF \g_stm_orcid_seq
  {
    \group_begin:
      \tex_let:D \thefootnote \relax \footnotetext
      {
        \raggedright
        \textsc{orcid}(s):\c_space_token
        \seq_use:Nn \g_stm_orcid_seq { ;~ }
      }
    \group_end:
  }
}
\ExplSyntaxOff

\begin{document}
\let\WriteBookmarks\relax

\shorttitle{Visual TCM Diagnostic System}
\shortauthors{Wang et~al.}

\title[mode=title]{Evidence-Based Intelligent Diagnostic and Therapeutic Visualization System with Large Language Models: Multi-Turn Interaction and Multimodal Treatment Plan Generation}

\author[1]{Yunhan Wang}
\credit{Conceptualization, Methodology, Software, Visualization, Writing -- original draft}

\author[1]{Yuda Wang}
\credit{Software, Validation, Visualization, Writing -- original draft}

\author[1,3]{Zhiying Tu}
\credit{Methodology, Data curation, Formal analysis, Writing -- review and editing}

\author[4]{Mingqiang Song}
\credit{Resources, Validation, Investigation, Writing -- review and editing}

\author[5]{Li Song}
\credit{Resources, Investigation, Validation, Writing -- review and editing}

\author[6]{Kun Li}
\credit{Resources, Project administration, Writing -- review and editing}

\author[1,3]{Dianhui Chu}
\credit{Supervision, Methodology, Project administration, Writing -- review and editing}

\author[1,2,3]{Bolin Zhang}
\cormark[1]
\ead{bolin@hit.edu.cn}
\credit{Conceptualization, Funding acquisition, Methodology, Supervision, Project administration, Writing -- review and editing}

\affiliation[1]{
  organization={Harbin Institute of Technology, Weihai},
  city={Weihai},
  state={Shandong},
  country={China}
}

\affiliation[2]{
  organization={Harbin Institute of Technology (Weihai) Qingdao Research Institute},
  city={Qingdao},
  state={Shandong},
  country={China}
}

\affiliation[3]{
  organization={Shandong Key Laboratory of Digital Service Computing Technology and Systems},
  country={China}
}

\affiliation[4]{
  organization={Weihai Municipal Hospital},
  city={Weihai},
  state={Shandong},
  country={China}
}

\affiliation[5]{
  organization={Shanghai Taizhu Technology Co., Ltd},
  city={Shanghai},
  country={China}
}

\affiliation[6]{
  organization={Tianjin Zhifu Qihuang Medical Technology Co., Ltd},
  city={Tianjin},
  country={China}
}

\cortext[cor1]{Corresponding author.}

\begin{abstract}
\textbf{Aim:} Existing AI-assisted traditional Chinese medicine diagnostic tools suffer from opaque reasoning processes, passive interaction, and limited treatment plan presentation. This study proposes a knowledge-enhanced visual diagnostic system to improve the transparency and interpretability of syndrome differentiation and treatment.

\textbf{Methods:} The system is built upon a Neo4j knowledge graph comprising 241 syndromes, 1,263 symptoms, and 2,485 relations. It incorporates a four-stage symptom matching pipeline (exact $\rightarrow$ semantic $\rightarrow$ fuzzy $\rightarrow$ large language model verification), an information gain--driven proactive questioning strategy optimized with genetic algorithms, and a multimodal treatment presentation integrating artificial intelligence-generated illustrations, three-dimensional meridian--acupoint models, and evidence-based literature.

\textbf{Results:} Knowledge graph constraints reduced non-standard outputs by 32\%. Case studies validated the effectiveness of the interactive workflow across patient self-assessment, clinician-assisted diagnosis, and traditional Chinese medicine education. Automated paired-comparison evaluation across 30 cases further demonstrated significant improvements in diagnostic trust (Cohen's $d=1.82$, $p<0.001$), reduced cognitive load (improvements in four of five dimensions), and higher credibility of evidence-based references (4.21 vs. 2.95).

\textbf{Conclusions:} The proposed system enhances the transparency of traditional Chinese medicine diagnostic reasoning and the interpretability of treatment plans through knowledge graph--driven visualization and multimodal interaction, offering a practical solution for trustworthy artificial intelligence-assisted traditional Chinese medicine applications.
\end{abstract}



\maketitle

\section{Introduction}\label{sec:introduction}
\input{sections/introduction}

\section{Methods}\label{sec:methods}
\input{sections/methods}

\section{Results}\label{sec:results}
\input{sections/results}

\section{Discussion}\label{sec:discussion}
\input{sections/discussion}

\section{Conclusion}\label{sec:conclusion}
\input{sections/conclusion}

\input{sections/declarations}

\printcredits

\bibliographystyle{unsrtnat}
\bibliography{references}

\end{document}

%% file: sections/introduction.tex

Traditional Chinese Medicine (TCM) diagnosis requires integrating inspection, auscultation-olfaction, inquiry, and palpation to map symptom combinations to standardized syndrome patterns~\cite{38}. Although AI-based diagnostic assistance has advanced rapidly in healthcare~\cite{7,8,14}, existing TCM AI tools face three critical problems: diagnostic reasoning remains opaque, functioning as a ``black box''~\cite{40,41}; interaction is predominantly passive, lacking multi-turn probing capability~\cite{11,36}; and treatment plans suffer from information overload due to monolithic text presentation~\cite{67,Baxter2025}. These deficiencies span knowledge representation, where the parametric knowledge of large language models (LLMs) is prone to hallucinations in medical domains~\cite{13,44,45}; interaction design, where single-turn question-answer modes lack progressive symptom clarification~\cite{1,19,DocCHA2025,61}; and information presentation, where diagnostic reasoning lacks visualization and treatment plans rely on text alone~\cite{24,30}. No existing system simultaneously integrates knowledge graph visualization, multi-turn interactive diagnosis, and multimodal treatment presentation.

\begin{figure*}[pos=t]
  \centering
  \includegraphics[width=\textwidth]{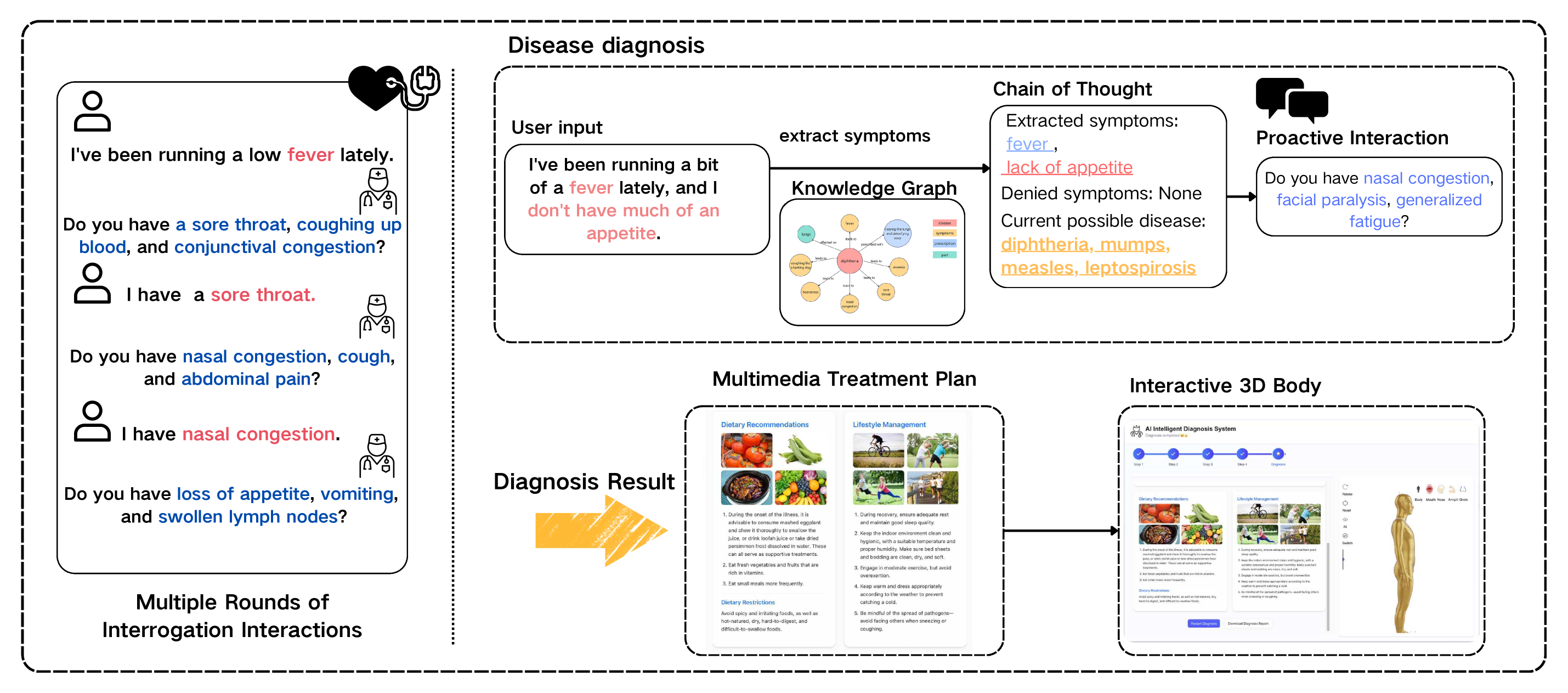}
  \caption{System overview. The left panel shows the multi-turn interactive consultation process: users describe symptoms in natural language, and the system progressively clarifies key differentiating information through knowledge-enhanced active questioning. The upper right panel illustrates the diagnostic reasoning process: four-layer progressive symptom matching combined with an information-gain-optimized questioning strategy performs candidate syndrome screening and convergence on a dynamic knowledge graph subgraph. The lower right panel shows multimodal diagnostic output: structured treatment plans, AI-generated treatment illustrations, evidence-based references, and an interactive three-dimensional acupoint model.}
  \label{fig:system-overview}
\end{figure*}

\FloatBarrier

Addressing these gaps, this paper investigates the following three research questions (RQs):

\begin{itemize}
  \item \textbf{RQ1:} How does the KG-grounded system perform relative to representative LLM baselines in syndrome differentiation accuracy and output standardization on real clinical records?

  That is, on 147 fine-grained syndrome labels drawn from real clinical records, how does a system whose candidate space is bounded by verified symptom--syndrome relations compare with unconstrained general-purpose and medical-domain LLMs? This is a system-level comparison and does not isolate the contribution of any single component.

  \item \textbf{RQ2:} How does knowledge graph visualization enhance the transparency of TCM diagnostic reasoning?

  Specifically, can progressive subgraph display help users understand the reasoning path from symptoms to syndrome types, thereby increasing trust in the system's diagnostic conclusions?

  \item \textbf{RQ3:} How does multimodal presentation improve patient comprehension of treatment plans?

  We evaluate two facets separately. \textbf{RQ3.1:} do AI-generated treatment illustrations and three-dimensional acupoint models lower the cognitive load of following a plan? \textbf{RQ3.2:} do structured evidence-based references raise the perceived credibility of the recommendations?
\end{itemize}

To answer these questions, we designed a knowledge-enhanced TCM visual diagnostic system (Figure~\ref{fig:system-overview}), built around a Neo4j knowledge graph containing 241 syndrome types, 1,263 symptoms, and 2,485 relationships. It combines four-layer progressive symptom matching with an information-gain-driven questioning strategy, enabling end-to-end visual interaction from symptom input through diagnosis to treatment planning. The main contributions are:

\begin{itemize}
  \item \textbf{C1: Evidence-based diagnostic visualization.} A progressive knowledge graph visualization method that makes TCM syndrome differentiation reasoning transparent and verifiable by dynamically updating symptom-syndrome subgraphs after each interaction round, allowing users to trace how diagnostic conclusions form.

  \item \textbf{C2: Multi-turn interactive diagnosis.} An interactive framework combining four-layer symptom matching (exact, vector semantic, fuzzy, and LLM verification) with information-gain-driven questioning to guide progressive symptom clarification within the knowledge-graph-defined syndrome space.

  \item \textbf{C3: Multimodal treatment presentation.} A framework integrating AI-generated treatment illustrations, a three-dimensional acupoint model, and structured evidence-based references, transforming text-only recommendations into an intuitive multimodal format.
\end{itemize}

The system targets three user groups: patients seeking preliminary self-assessment through guided visual interaction, clinicians using it as a ``second opinion'' tool with full reasoning path transparency, and TCM students learning syndrome differentiation through interactive diagnostic case replay.

A comprehensive review of related work, including comparisons with representative systems, is provided in Supplementary Information Section~S1.

%% file: sections/methods.tex

\subsection{Design Challenges}\label{sec:challenges}

Before designing the system, we identified core design challenges facing AI-assisted TCM diagnostic systems through three channels: (1) a systematic review of existing AI-assisted diagnosis systems and medical visualization literature; (2) informal discussions with two TCM practitioners; and (3) an analysis of the limitations of current AI chat tools in clinical scenarios.
These challenges, distilled from literature and practitioner input, guided the system design decisions below.

\begin{figure*}[pos=t]
  \centering
  \includegraphics[width=\textwidth]{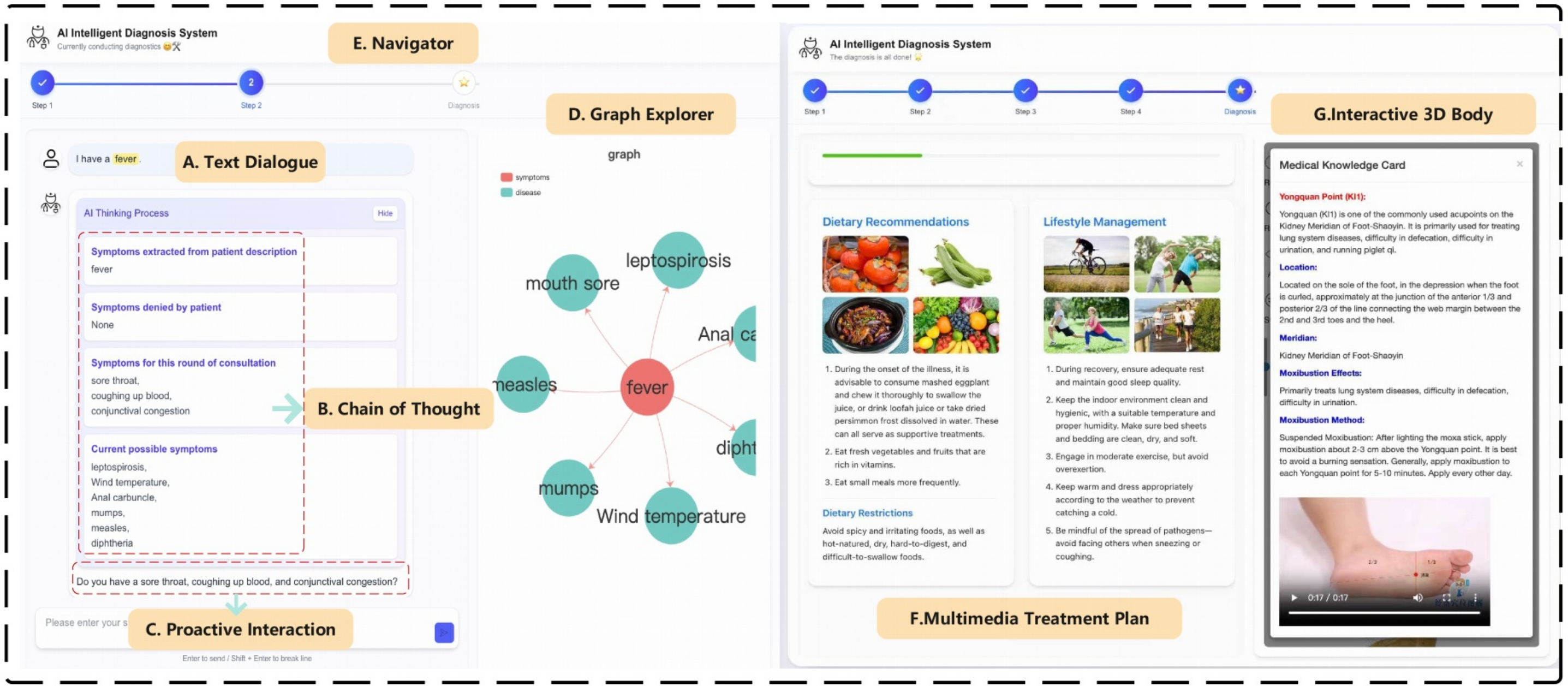}
  \caption{System interface overview. The interface integrates seven core functional areas: (A)~Natural language conversational interaction area, where users describe symptoms as text and engage in multi-round consultation with the system; (B)~Chain-of-thought reasoning display area, showing the system's step-by-step diagnostic reasoning in real time; (C)~Proactive interaction confirmation area, with system-generated symptom confirmation and follow-up prompts; (D)~Knowledge graph browsing area, displaying the case-specific symptom-syndrome association subgraph; (E)~Function navigation menu, providing access to diagnostic history, settings, and other auxiliary functions; (F)~Multimodal treatment plan area, integrating text recommendations, AI-generated illustrations, and references; (G)~3D meridian-acupoint model, an interactive bronze figure model for precise localization of recommended acupoints.}
  \label{fig:system-interface}
\end{figure*}

\paragraph{DC1: Prior Dependency and Hallucination Risk}
AI-assisted diagnosis faces two intertwined ``black-box'' risks: opaque reasoning paths that users cannot trace~\cite{40,41} and LLM hallucinations that fabricate rationales or associations~\cite{44,45}. We address both by grounding interaction in a progressively updated, case-specific KG subgraph: it provides structured visualization for interpretability~\cite{24,35} while also constraining generation within verified symptom--syndrome relations~\cite{25,64}. This constraint depends on KG coverage, label alignment, and prompt strategy.

\paragraph{DC2: Ambiguous Expressions and Syndrome Space}
Colloquial and imprecise symptom descriptions can be misaligned with KG entities~\cite{65}, and early matching errors can cascade into diagnostic drift. Meanwhile, effective medical consultation requires multi-turn proactive clarification when user information is incomplete~\cite{1,19,DocCHA2025,61}, yet many existing systems lack such capability. We therefore combine robust, progressive symptom matching with information gain-driven, multi-round follow-up~\cite{11,36,12}, using user confirmations/denials as implicit feedback to iteratively narrow the KG-defined candidate syndrome space.

\paragraph{DC3: Cognitive Threshold and Literacy Disparities}
Text-only treatment recommendations can overwhelm non-expert users, especially for unfamiliar concepts such as formula composition and acupoint localization~\cite{67,Baxter2025}. This is exacerbated by heterogeneous audiences (patients, physicians, and students) with different literacy and evidence needs~\cite{7,14}. We therefore use a multimodal presentation framework that integrates text with AI-generated illustrations and interactive 3D acupoint visualization~\cite{30,69}, plus structured references, so users can access information at appropriate depth.

\subsection{System Design}\label{sec:design}

We describe the three-tier architecture and four functional modules (M1--M4), and summarize the diagnosis--treatment dual-chain visualization spanning the workflow.

\subsubsection{System Overview}\label{sec:design-overview}

\begin{figure}[pos=tbp]
  \centering
  \begin{subfigure}[b]{\linewidth}
    \centering
    \includegraphics[width=\linewidth]{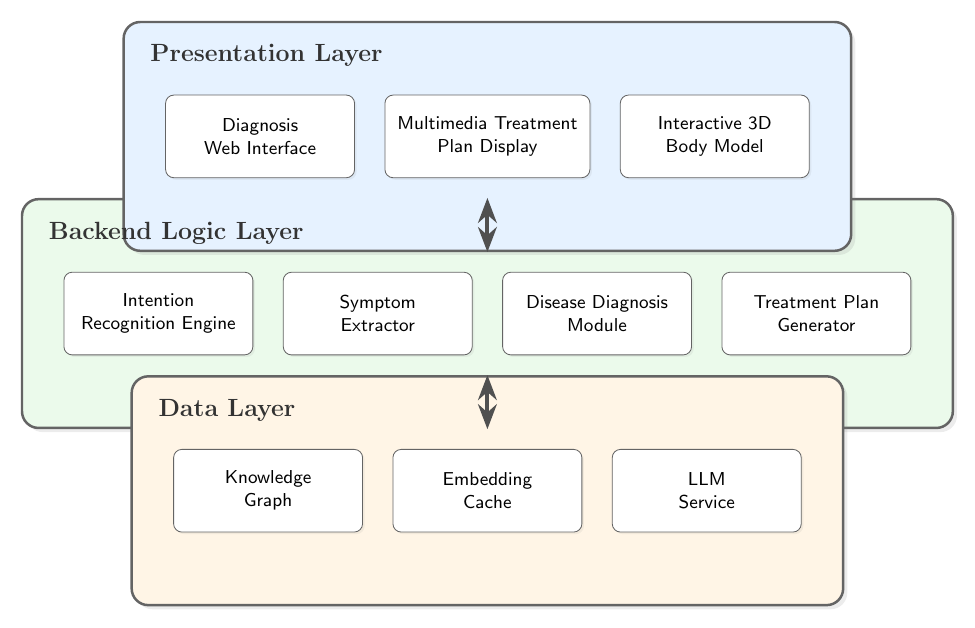}
    \caption{}
    \label{fig:architecture}
  \end{subfigure}

  \begin{subfigure}[b]{\linewidth}
    \centering
    \includegraphics[width=\linewidth]{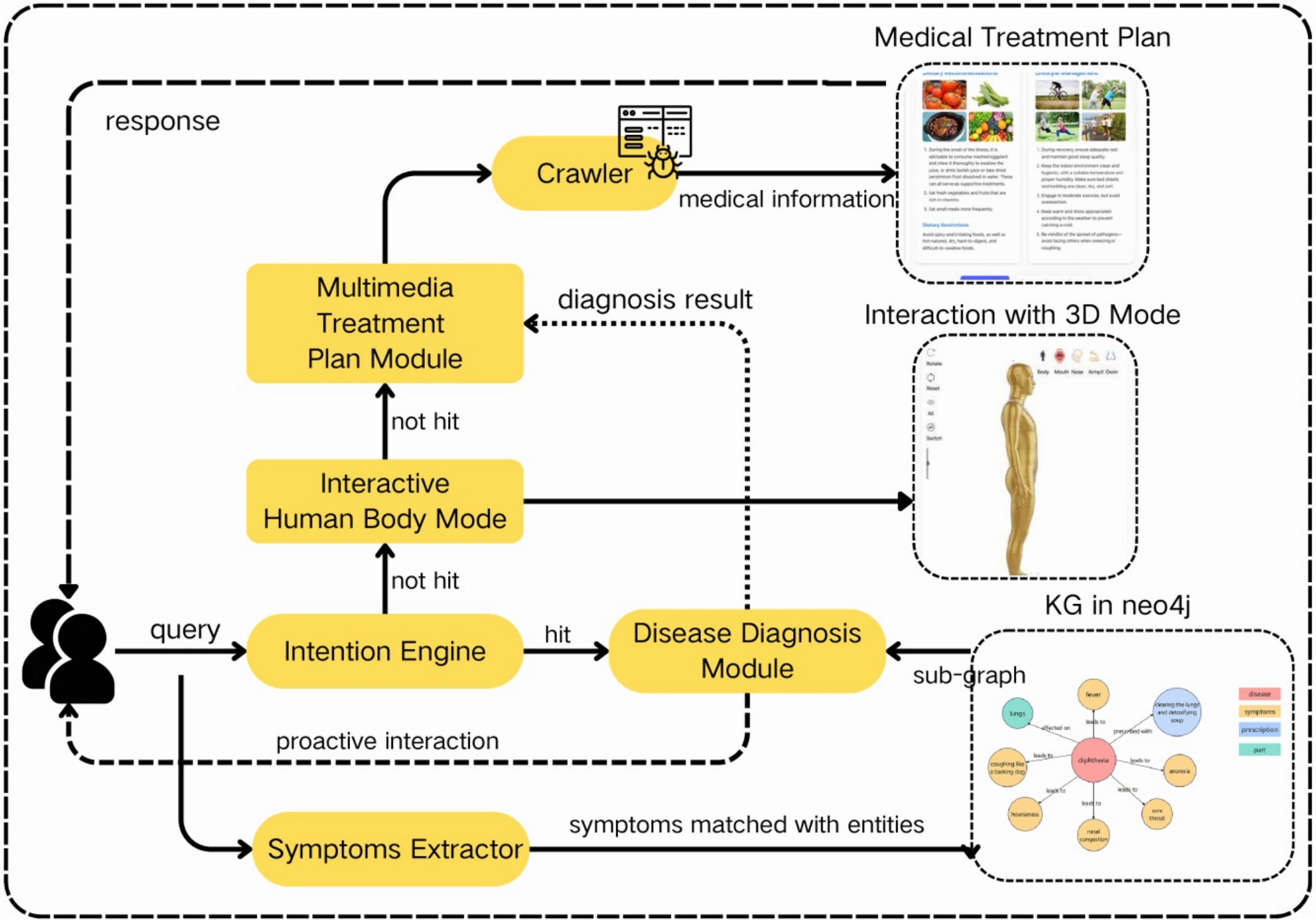}
    \caption{}
    \label{fig:workflow-4stage}
  \end{subfigure}
  \caption{System architecture and interaction workflow. (a)~Three-tier architecture: the front-end presentation layer adopts a left-right split-screen layout, with conversational interaction on the left and knowledge visualization on the right; the back-end logic layer integrates core modules including symptom extraction, multi-layer matching, and LLM interaction; the data layer centers on a graph database supporting KG storage and querying. (b)~Four-module interaction workflow, covering symptom collection, progressive follow-up, syndrome differentiation diagnosis, and treatment plan presentation; the upper portion shows the core interaction logic and data flow of each module, and the lower portion shows the knowledge graph and data service layer supporting the entire process.}
  \label{fig:arch-workflow}
\end{figure}

The system presents a split-screen interface (Figure~\ref{fig:system-interface}) backed by a three-tier architecture (Figure~\ref{fig:architecture}). The front-end supports conversational consultation and interactive visualization; the back-end orchestrates symptom extraction, KG-grounded follow-up, and LLM interaction; and the data layer stores the KG and references for efficient retrieval. Standardized APIs enable modular iteration across tiers.

\subsubsection{M1: Knowledge Graph Construction}\label{sec:design-m1}

Users begin by describing symptoms in the conversational panel (Figure~\ref{fig:workflow-4stage}). The system extracts candidate symptom entities, highlights matches, and asks users to confirm or deny them, providing implicit feedback that helps refine symptom alignment~\cite{11,Wei2018}. In parallel, it renders an initial case-specific KG subgraph from confirmed symptoms to candidate syndrome types, making the starting hypothesis space visible~\cite{24}.

\subsubsection{M2: Knowledge-guided Proactive Asking}\label{sec:design-m2}

\begin{figure}[pos=tbp]
  \centering
  \includegraphics[width=\linewidth]{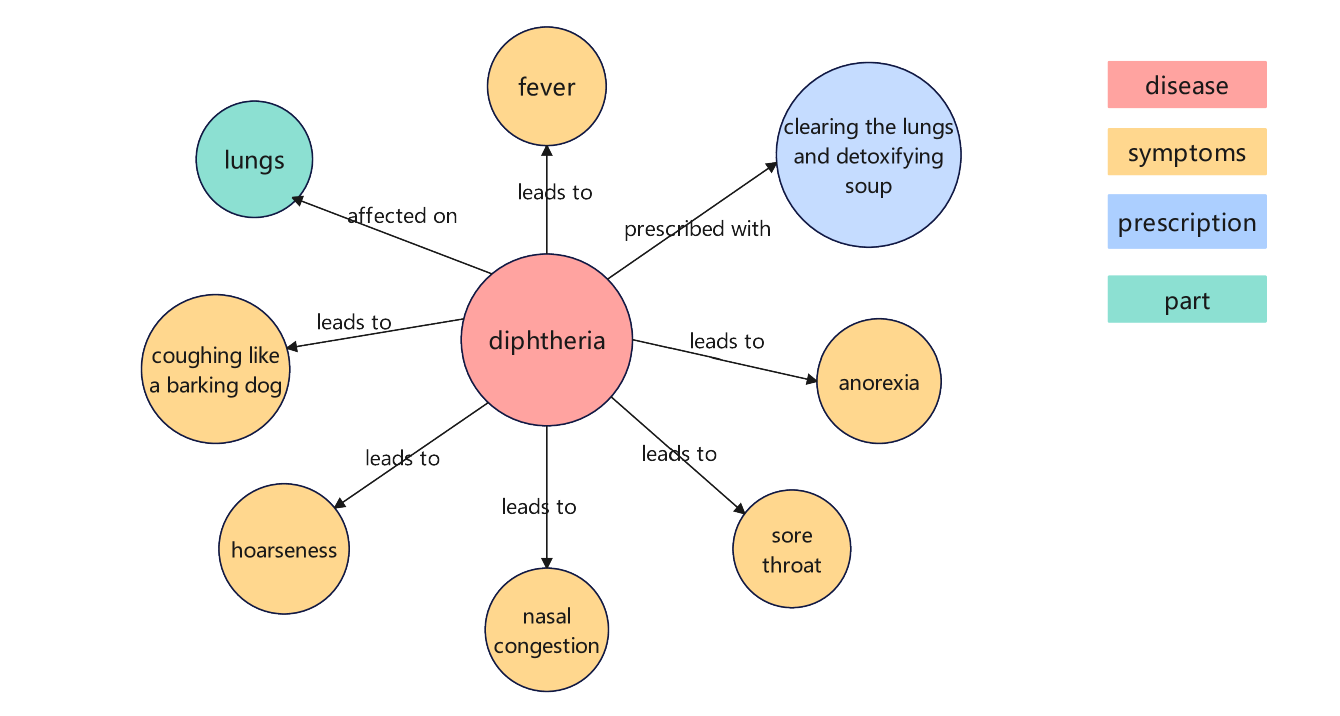}
  \caption{Knowledge graph visualization. The figure shows a case-specific knowledge subgraph constructed around the diphtheria syndrome type, with the target syndrome type as the central node surrounded by associated symptom nodes and treatment method nodes. Edges represent semantic relations.}
  \label{fig:kg-visualization}
\end{figure}

\begin{figure*}[pos=tbp]
  \centering
  \includegraphics[width=\linewidth]{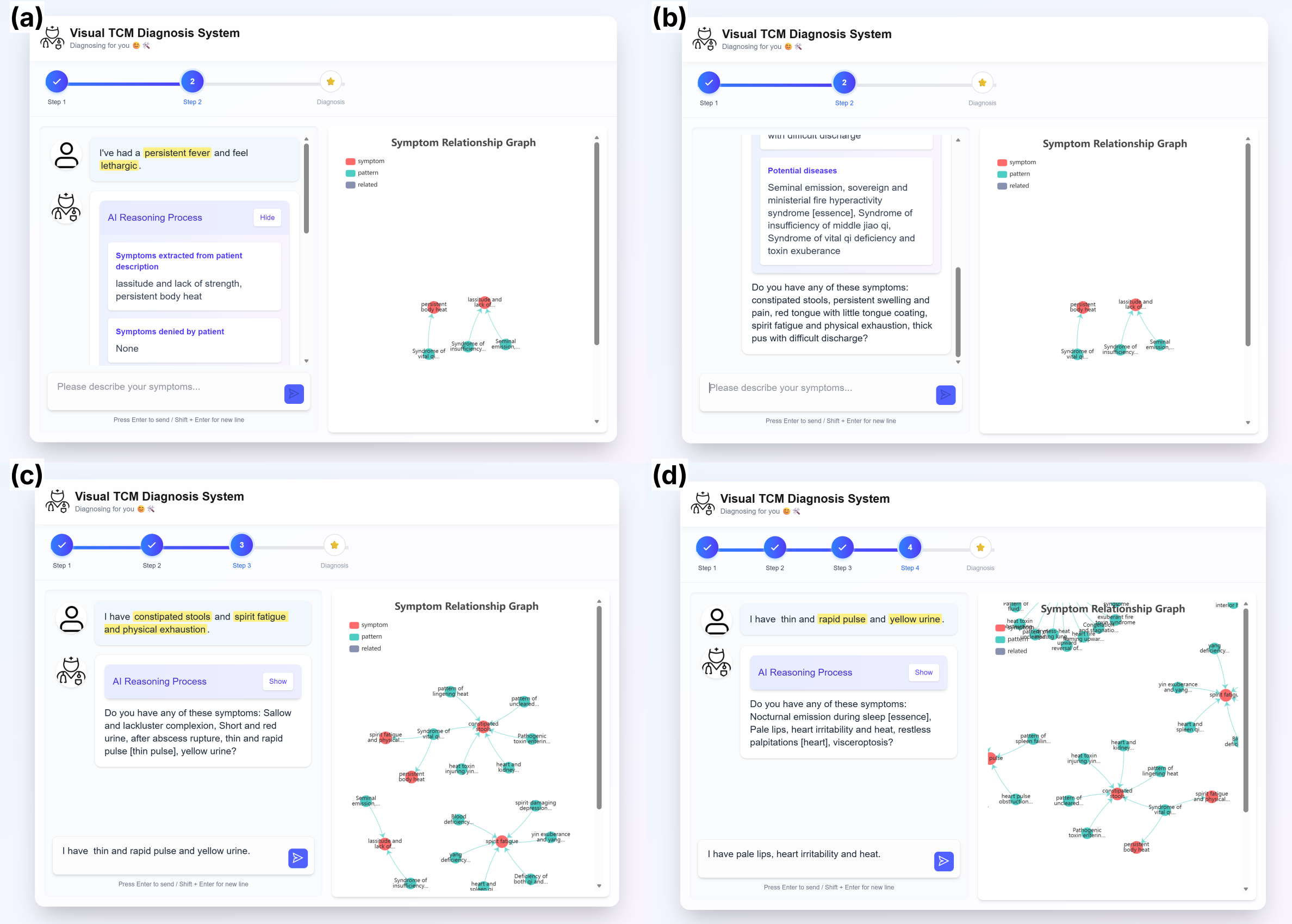}
  \caption{Multi-round follow-up consultation interaction process. (a)~Symptom extraction and KG construction: the user inputs symptoms (e.g., fever, fatigue), the system extracts keywords and renders the symptom relationship graph on the right; (b)~Intelligent follow-up: the system generates targeted follow-up questions based on association relationships in the KG; (c)~Multi-round dialogue continuation: the user and system engage in multiple rounds of symptom confirmation and supplementation; (d)~Follow-up completion: the KG expands further, and after the user supplements symptoms such as pale lips, candidate syndrome types converge.}
  \label{fig:multi-round-followup}
\end{figure*}

Given the current candidate syndrome space, the system proactively asks follow-up questions selected to maximize information gain and reduce uncertainty~\cite{11,12}. After each response, confirmed and negated symptoms update the candidate space and the case-specific KG subgraph, which refreshes in real time (Figure~\ref{fig:kg-visualization}). Figure~\ref{fig:multi-round-followup} shows an example multi-round session. The progressive KG both bounds generation to KG-defined associations and provides a verification surface for monitoring evidence accumulation~\cite{25,64}.

\subsubsection{M3: In-depth Consultation and Syndrome Differentiation Diagnosis}\label{sec:design-m3}

\begin{figure*}[pos=tbp]
  \centering
  \includegraphics[width=\linewidth]{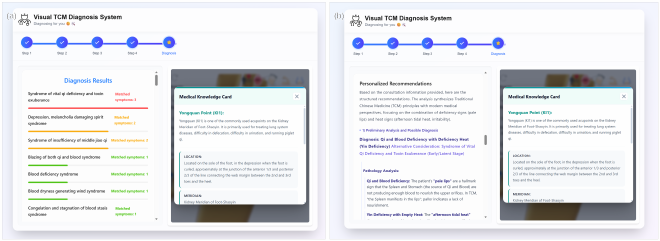}
  \caption{Diagnostic results and treatment plan presentation. The left panel shows candidate syndrome type diagnostic cards ranked by matching degree; the right panel shows the personalized treatment recommendation overview generated for the confirmed syndrome type.}
  \label{fig:diagnosis-treatment}
\end{figure*}

The system performs final differentiation among remaining candidate syndromes and may ask targeted questions about discriminative signs~\cite{DocCHA2025,BRAgent2022}. Results are presented as ranked diagnostic cards with matched symptom evidence and a visual matching degree (Figure~\ref{fig:diagnosis-treatment}), supporting progressive disclosure while preserving KG-based traceability~\cite{66,11}.

\subsubsection{M4: Diagnostic Results and Multimodal Treatment Recommendations}\label{sec:design-m4}

\begin{figure}[pos=tbp]
  \centering
  \begin{subfigure}[b]{\linewidth}
    \centering
    \includegraphics[width=\linewidth]{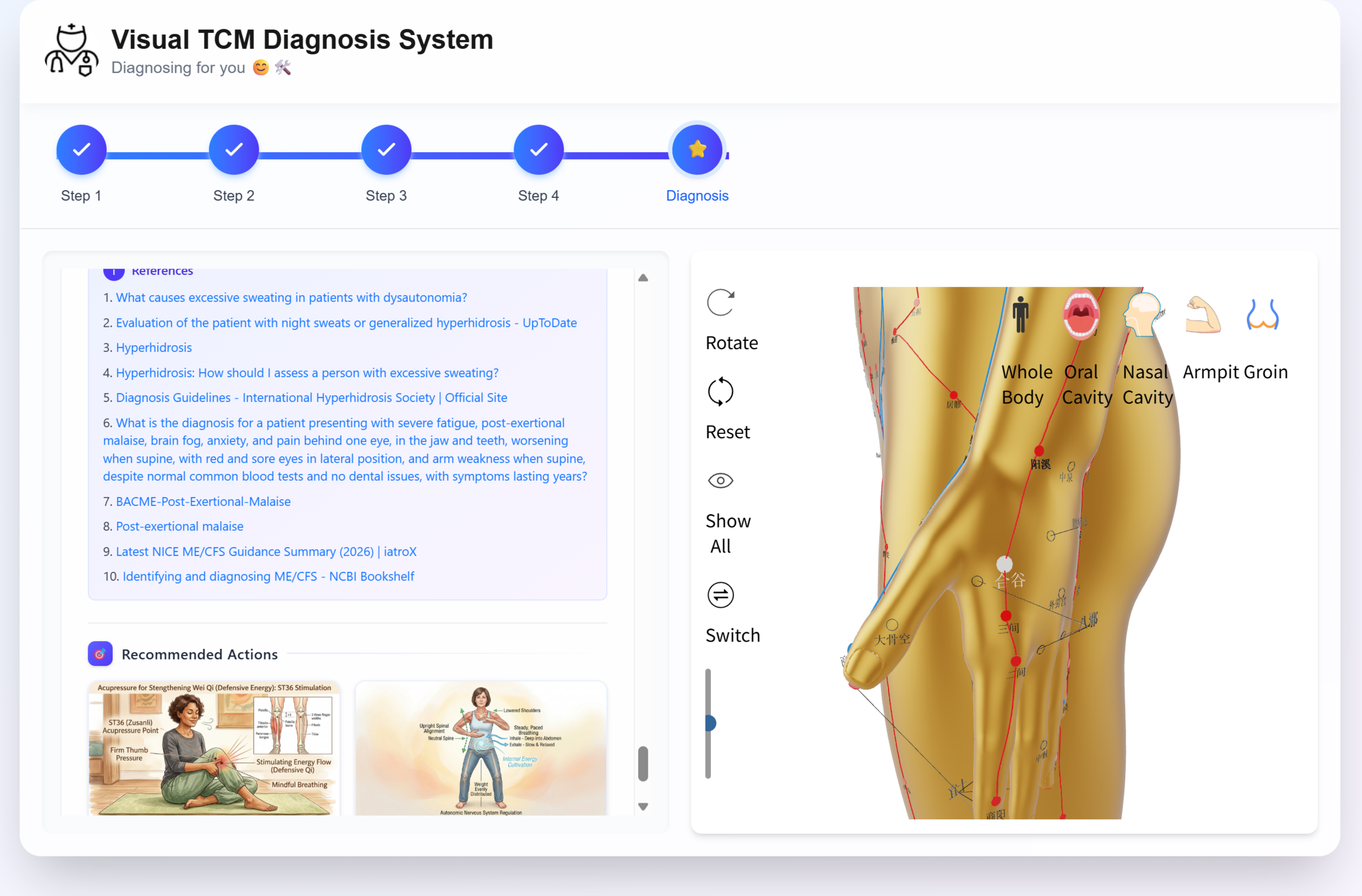}
    \caption{}
    \label{fig:ai-illustrations}
  \end{subfigure}

  \begin{subfigure}[b]{\linewidth}
    \centering
    \includegraphics[width=\linewidth]{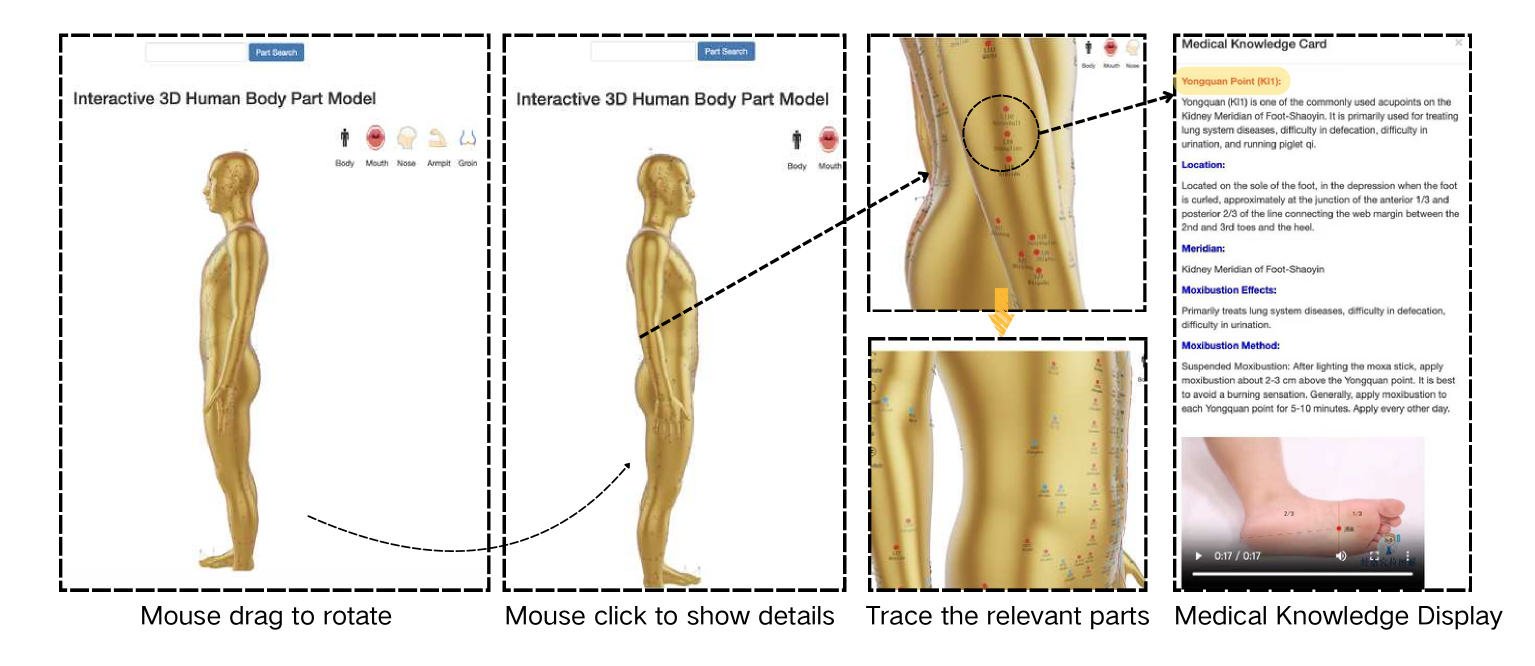}
    \caption{}
    \label{fig:3d-acupoint}
  \end{subfigure}
  \caption{Multimodal treatment presentation. (a)~Treatment recommendation interface: the upper-left panel lists evidence-linked references supporting the recommendation, and the lower-left panel presents AI-generated recommended-action illustrations. (b)~3D acupoint interactive display: a rotatable bronze-figure meridian model for locating recommended acupoints and viewing associated massage guidance.}
  \label{fig:treatment-presentation}
\end{figure}

After diagnosis confirmation, the system generates a personalized treatment plan grounded in KG associations, covering formulas, dietary therapy, lifestyle guidance, and acupoint massage, with evidence links for verifiability~\cite{MedCite2025,14}. In the treatment view, the left side combines an evidence-linked reference list with AI-generated recommended-action illustrations (Figure~\ref{fig:ai-illustrations}), while a separate 3D meridian--acupoint view remains available for massage guidance (Figure~\ref{fig:3d-acupoint})~\cite{63}. This keeps the main text focused on the two presentation aids most relevant to comprehension, without unpacking every interface submodule~\cite{30}.

\subsubsection{Diagnosis-Treatment Dual-Chain Visualization}\label{sec:design-dualchain}

We conceptualize the workflow as a \textbf{diagnosis--treatment dual-chain visualization} architecture. The diagnostic chain (M1--M3) externalizes the evolving symptom--syndrome hypothesis space through progressive KG subgraphs and multi-round follow-up, enabling transparency and verification. The treatment chain (M4) transforms the confirmed syndrome type into an accessible, multimodal plan. Both chains share the KG as a unified backbone that links evidence accumulation during diagnosis to recommendation presentation during treatment.

\subsection{Implementation}\label{sec:implementation}

The system is implemented as a three-tier web architecture with a separated front-end, back-end, and graph-data layer. The front-end uses Next.js~15 and React~19 to provide the interactive diagnostic and treatment interface. The back-end uses Flask REST APIs to host diagnostic reasoning, symptom matching, follow-up question selection, treatment recommendation, and evaluation-related services. The data layer uses Neo4j to store the TCM knowledge graph, which contains 241 syndrome types, 1,263 symptom entities, and 2,485 symptom--syndrome relationships. The LLM service is accessed through Gemini~3.1~Pro (Google), and semantic matching uses BGE-M3 embeddings. The system runs on Python~3.11 and Node.js~18+ and supports Docker-based deployment.

\paragraph{Four-layer symptom matching}
To align free-text symptom descriptions with standardized KG symptom entities, the system applies a progressive four-layer matching pipeline. Layer~1 performs case-insensitive exact matching against KG symptom labels. Layer~2 applies semantic vector matching with BGE-M3 embeddings. For a user symptom mention $\nu$ and a KG symptom entity $\xi \in \Omega$, cosine similarity is computed as:
\begin{equation}\label{eq:main-sim}
\mathrm{sim}(\nu, \xi) =
\frac{\mathbf{e}(\nu) \cdot \mathbf{e}(\xi)}
{\|\mathbf{e}(\nu)\| \cdot \|\mathbf{e}(\xi)\|},
\end{equation}
where $\mathbf{e}(\cdot)$ denotes the embedding function. The best semantic match is accepted when its similarity exceeds threshold $\delta$. Layer~3 applies fuzzy matching with the rapidfuzz \texttt{token\_set\_ratio} algorithm to handle spelling variants and abbreviated expressions, using threshold $\tau$. Layer~4 invokes the LLM only for unresolved or ambiguous cases and requires a structured JSON verification output.

Let $s^\ast$ and $f^\ast$ denote the best semantic and fuzzy candidates, respectively. The resulting cascading matching function is:
\begin{equation}\label{eq:main-match}
\mathrm{match}(\nu) = \begin{cases}
\xi & \text{if } \nu = \xi \text{ by exact match}, \\
s^\ast & \text{if } \mathrm{sim}(\nu,s^\ast) \geq \delta, \\
f^\ast & \text{if } \mathrm{fuzz}(\nu,f^\ast) \geq \tau, \\
\mathrm{LLM}(\nu,\Omega) & \text{otherwise}.
\end{cases}
\end{equation}
This design resolves common cases through efficient symbolic, vector, and fuzzy operations, while reserving the higher-cost LLM fallback for cases where deterministic matching is insufficient.

\paragraph{Information-gain follow-up selection}
At each interaction round, the system selects a discriminative symptom subset $Q_t$ from the unasked symptom space to reduce uncertainty over the current candidate syndrome set $\mathcal{Y}_t$. We use a weighted utility function:
\begin{equation}\label{eq:main-utility}
U(Q_t) =
\alpha \cdot \mathrm{IG}(Q_t;\mathcal{Y}_t)
- \beta \cdot R(Q_t)
+ \gamma \cdot C(Q_t),
\end{equation}
where $\mathrm{IG}$ measures entropy reduction in the candidate syndrome set after asking $Q_t$, $R(Q_t)$ penalizes highly correlated symptom questions, and $C(Q_t)$ encourages symptoms that cover discriminative features across candidate syndrome types. The coefficients $\alpha$, $\beta$, and $\gamma$ control the relative weights. Because exhaustive search over symptom subsets is computationally infeasible when the candidate symptom space is large, the system uses a genetic algorithm for approximate optimization, with $U(Q_t)$ as the fitness function.

\paragraph{Engineering and visualization implementation}
Several implementation choices were made to support responsive interaction. At application startup, syndrome--symptom mappings are loaded from Neo4j into a KG warm-up cache, reducing repeated graph queries during diagnosis. BGE-M3 vectors for KG symptom names are precomputed and persisted with a SHA-256 vocabulary hash as the cache invalidation key. Batch embedding requests use \texttt{ThreadPoolExecutor} for parallel computation. LLM structured outputs are handled through a three-stage JSON parsing and repair pipeline: direct parsing, regular-expression extraction from Markdown code fences, and LLM self-repair/regeneration. The front-end uses skeleton loading states to keep the interface stable during API calls.

The KG visualization is implemented with an ECharts force-directed graph embedded in the right panel of the diagnostic interface. The visualization refreshes when confirmed symptoms or candidate syndrome types change, using red symptom nodes, green syndrome nodes, and highlighted selected nodes to distinguish entity roles. During treatment presentation, a 3D acupoint model supports rotation and zoom interactions for locating recommended acupoints, while symptom keywords in the chat interface are highlighted to make extracted symptom evidence visible to users. Additional implementation details are provided in Supplementary Information Section~S2.

\subsection{Automated Evaluation Methodology}\label{sec:eval-methodology}

To evaluate key design decisions reproducibly, we used a mixed strategy: a quantitative accuracy evaluation for syndrome differentiation (RQ1) and an \textbf{LLM-as-a-Judge} framework for paired, structured assessments of interaction and presentation choices (RQ2, RQ3.1, RQ3.2). RQ1 evaluates Top-1 syndrome differentiation accuracy on a de-identified CEMRs dataset containing over 2,000 records and 147 fine-grained syndrome labels. For the LLM-as-a-Judge experiments, cases were drawn from a separate de-identified set of 160 TCM electronic medical records. We used stratified random sampling by syndrome type (random seed $=42$), allocated quotas proportionally to syndrome frequency while ensuring at least one case per sampled syndrome type, and selected $N=30$ cases covering no fewer than five syndrome types. RQ2 and RQ3.1 used all 30 cases; RQ3.2 used 25 cases after excluding records for which the treatment recommendation API did not return valid references. Full details of case selection and statistical analysis are given in Supplementary Information Section~S4.

Claude Sonnet~4.5 (Anthropic) served as the automated judge model. To reduce same-source bias, the judge model was selected from a different vendor than the system model and patient simulator, which were from the Google model family. All judging calls used deterministic decoding (\texttt{temperature=0}). For each evaluation dimension, the judge assigned a 1--5 Likert score, where 1 indicates very poor performance and 5 indicates very good performance, and returned structured JSON containing scores and brief reasoning. Outputs were validated through the same three-stage JSON handling strategy used by the system: direct parsing, regular-expression extraction, and pattern-based fallback. Missing dimensions triggered up to two retries; if retries failed, the default minimum score was assigned for the missing dimension.

Given the paired design, small sample size ($n \leq 30$), and ordinal score data, between-condition differences were tested using the two-sided Wilcoxon signed-rank test with significance level $\alpha=0.05$. Benjamini--Hochberg false discovery rate correction was applied across dimensions for each research question. Effect sizes were reported as paired-sample Cohen's $d$ for RQ2 and RQ3.2. For RQ3.1, because reverse-scored cognitive-load dimensions produced many zero differences, we used rank-biserial $r$, computed from the Wilcoxon statistic as $r = 1 - 2W/[n(n+1)]$.

\paragraph{Evaluation Framework Overview}
We address four evaluation questions aligned with the three research questions. \textbf{RQ1} compares the KG-grounded system with representative LLM baselines on Top-1 syndrome differentiation accuracy and output standardization; it is a system-level comparison, not an ablation. \textbf{RQ2} tests whether KG visualization improves diagnostic trust by comparing the same consultation with and without KG augmentation. A Gemini~3~Flash patient simulator completed up to five diagnostic rounds for each case, and the judge then evaluated diagnostic transparency, evidence traceability, reasoning confidence, information completeness, and overall trust from the patient perspective. \textbf{RQ3.1} tests whether multimodal treatment plans reduce cognitive load by comparing a plain-text plan (Plan~A) with a structured multimodal plan (Plan~B) containing step-by-step guidance, AI-generated ingredient illustrations, and 3D acupoint-model descriptions. The judge used the persona of a non-expert user with limited medical background and assessed language comprehensibility, preparation clarity, acupoint localizability, action execution confidence, and overall cognitive load; the cognitive-load item was reverse-scored so that higher values consistently indicate better user experience. \textbf{RQ3.2} tests whether reference display improves evidence credibility by comparing treatment recommendations with and without references. In the reference condition, the judge received titles, URLs, abstract snippets, and scraped page-content excerpts truncated to the first 2,000 characters, and assessed source credibility, evidence verifiability, information confidence, comprehension support, and overall evidence quality. For all paired A/B judging experiments, condition order was randomized deterministically by case ID. Full prompts for the patient simulator and all judge conditions are documented in Supplementary Information Section~S5.

%% file: sections/results.tex

\subsection{Evaluation Strategy}\label{sec:eval-strategy}

This paper focuses on overall system design and interaction experience rather than algorithmic peak performance.
We adopted a four-pronged evaluation strategy:
(1)~\textbf{Technical accuracy verification}: quantitative syndrome differentiation accuracy on a clinical dataset, compared against mainstream large language models (Section~\ref{sec:eval-accuracy});
(2)~\textbf{Case study}: a complete diagnostic workflow demonstrating how design decisions operate together (Supplementary Information Section~S3);
(3)~\textbf{System capability assessment}: response time and rendering performance analysis (Section~\ref{sec:eval-performance});
(4)~\textbf{LLM-as-a-Judge automated evaluation}: paired comparison evaluations assessing trust, cognitive load, and reference credibility to quantify each design decision's effect (Sections~\ref{sec:eval-rq2-trust}--\ref{sec:eval-rq3c-refs}).
All evaluation data were de-identified, and no real patient interactions were involved.

\subsection{RQ1: Comparative Performance on Syndrome Differentiation}\label{sec:eval-accuracy}

\paragraph{Dataset}
We used the CEMRs dataset containing over 2,000 de-identified Chinese electronic medical records covering 147 unique syndrome pattern labels drawn from real clinical settings.

\paragraph{Comparison Methods}
To position the KG-grounded system against current practice, we compared against two baselines:
(1)~DeepSeek R1~\cite{58}, a general-purpose large language model with strong reasoning capabilities;
(2)~HuatuoGPT-Vision~\cite{62}, a medical domain-specific model fine-tuned on medical corpora.
This is a system-level comparison against two different models; it is not a component-wise ablation, so the observed differences cannot be attributed to the knowledge graph constraint alone, and no claim of comprehensive superiority is made.

\paragraph{Results and Analysis}
Table~\ref{tab:accuracy-comparison} presents the Top-1 syndrome differentiation accuracy on the CEMRs dataset.

\input{paper_tables/accuracy_comparison}

Several contextual factors inform these results.
The dataset contains 147 fine-grained syndrome labels (far exceeding the 10--20 coarse-grained categories in common benchmarks), and cases come from real clinical records with inherent noise such as colloquial language and ambiguity.

After introducing knowledge graph constraints, non-standardized outputs decreased by approximately 32\% compared to unconstrained LLMs, indicating that the knowledge graph constrains the generation space and reduces non-standard outputs, though hallucination cannot be eliminated entirely.

\paragraph{Reproducibility Statement}
Due to privacy restrictions, the CEMRs dataset cannot be publicly released. We provide 160 de-identified sample cases (\texttt{public\_dataset.json}) with complete prompt templates for independent validation.

\subsection{RQ2: Effect of Knowledge Graph Enhancement on Trust}\label{sec:eval-rq2-trust}

Using the LLM-as-a-Judge approach, we generated diagnostic results with and without knowledge graph enhancement for 30 CEMRs cases.
A large language model simulating target users performed paired comparison evaluations across five trust dimensions: diagnostic transparency, evidence traceability, reasoning confidence, information completeness, and trust level.
Each dimension was tested using the Wilcoxon signed-rank test with Benjamini--Hochberg correction.
Table~\ref{tab:rq2-trust} summarizes the paired trust evaluation results.

\input{paper_tables/rq2_trust_table}

Knowledge graph enhancement produced significant improvements across all five dimensions (all $p_{\mathrm{adj}}<0.001$), with an overall Cohen's $d=1.82$.
\textbf{Information completeness} showed the largest effect ($d=2.18$), indicating that structured knowledge graph presentation substantially enhanced users' perception of diagnostic information coverage.
\textbf{Evidence traceability} followed ($d=1.73$), reflecting clearer reasoning paths through graph visualization.
\textbf{Reasoning confidence} had the smallest effect ($d=0.87$), still within the large effect range, possibly because this dimension involves deeper cognitive judgment.

\subsection{RQ3.1: Effect of Multimodal Plans on Cognitive Load}\label{sec:eval-rq3a-cognitive}

We generated text-only plans (Plan~A) and multimodal plans (Plan~B, with 3D acupoint models and AI-generated illustrations) for 30 cases, evaluated by a large language model simulating non-specialist users.
Five cognitive dimensions were assessed; cognitive load was reverse-scored (higher~=~lower load, positive~$\Delta$~=~Plan~B advantage).
Table~\ref{tab:rq3-cognitive} reports the cognitive load comparison.

\input{paper_tables/rq3_cognitive_load}

Four of five dimensions reached statistical significance ($p_{\mathrm{adj}}<0.001$).
\textbf{Acupoint locatability} showed the most pronounced improvement ($\Delta=+1.87$, $r=1.00$), directly reflecting the 3D bronze figure model's advantage for spatial localization.
\textbf{Action confidence} ($\Delta=+0.70$, $r=0.91$) and \textbf{language comprehensibility} ($\Delta=+0.50$, $r=0.88$) also improved significantly.
\textbf{Preparation clarity} showed a positive trend ($\Delta=+0.27$) but did not reach significance after correction ($p_{\mathrm{adj}}=0.053$).
Overall cognitive load improved significantly ($\Delta=+0.60$, $r=0.90$), confirming that multimodal plans effectively reduced cognitive load.

\subsection{RQ3.2: Effect of References on Evidence Credibility}\label{sec:eval-rq3c-refs}

Paired comparison evaluations of plans with and without references were conducted across 30 cases; five were excluded due to invalid reference URLs, yielding 25 valid cases.
Five dimensions were assessed: source credibility, evidence verifiability, information confidence, comprehension support, and overall evidence quality.
Table~\ref{tab:rq3-references} reports the reference credibility evaluation.

\input{paper_tables/rq3_references}

Plans with references scored 4.21 overall, significantly higher than 2.95 without ($\Delta=+1.27$); four of five dimensions reached significance ($p_{\mathrm{adj}}<0.001$).
\textbf{Evidence verifiability} had the largest effect (Cohen's $d=1.98$), indicating that references transformed recommendations from unverifiable assertions into traceable evidence-based information.
\textbf{Overall evidence quality} ($d=1.55$) and \textbf{source credibility} ($d=1.52$) also showed large improvements.
\textbf{Comprehension support} did not reach significance ($p_{\mathrm{adj}}=0.317$, $d=-0.20$), aligning with expectations: references primarily enhance traceability and credibility rather than plan comprehensibility.

Beyond these quantitative evaluations, we conducted illustrative case studies across representative clinical scenarios to validate the system's end-to-end diagnostic workflow, covering patient self-assessment, physician-assisted diagnosis, and TCM education contexts.
Full case study results are presented in the Supplementary Information (Section~S3).

\subsection{System Performance}\label{sec:eval-performance}

At the engineering level, the system demonstrated good response efficiency: the average response time per interaction round was under 30~seconds,
the front-end ECharts knowledge graph visualization update rendered within 200~ms, and the overall interaction fluency met real-time dialogue requirements.
Limitations of the current system, covering knowledge graph syndrome coverage, evaluation scale, LLM output reliability, dataset external validation, and cross-language applicability, are discussed in Section~\ref{sec:limitations}.

%% file: paper_tables/accuracy_comparison.tex
\begin{table}[pos=tbp]
  \centering
  \footnotesize
  \caption{RQ1: Syndrome differentiation accuracy on the CEMRs dataset (over 2,000 records, 147 syndrome labels).}
  \label{tab:accuracy-comparison}
  \begin{tabular}{lc}
    \toprule
    Method & Top-1 accuracy \\
    \midrule
    Our system & $47.11\%$ \\
    DeepSeek R1 & $19.76\%$ \\
    HuatuoGPT-Vision & $11.23\%$ \\
    \bottomrule
  \end{tabular}
\end{table}

%% file: paper_tables/rq2_trust_table.tex
\begin{table*}[pos=tbp]
  \centering
  \footnotesize
  \caption{RQ2: Impact of knowledge graph visualization on diagnostic trust (LLM-as-a-Judge, $N=30$).}
  \label{tab:rq2-trust}
  \begin{threeparttable}
    \begin{tabular}{lccccc}
      \toprule
      Dimension & With KG & Without KG & $p$ & $p_{\mathrm{adj}}$ & Cohen's $d$ \\
      \midrule
      Diagnostic transparency & $2.87 \pm 0.63$ & $2.17 \pm 0.38$ & $<0.001$ & $<0.001$ & $1.35$ \\
      Evidence traceability & $3.37 \pm 0.61$ & $2.40 \pm 0.50$ & $<0.001$ & $<0.001$ & $1.73$ \\
      Reasoning confidence & $2.50 \pm 0.57$ & $2.10 \pm 0.31$ & $<0.001$ & $<0.001$ & $0.87$ \\
      Information completeness & $3.40 \pm 0.50$ & $2.33 \pm 0.48$ & $<0.001$ & $<0.001$ & $2.18$ \\
      Trust level & $2.70 \pm 0.60$ & $2.10 \pm 0.31$ & $<0.001$ & $<0.001$ & $1.27$ \\
      \midrule
      Overall & $2.97 \pm 0.51$ & $2.22 \pm 0.28$ & $<0.001$ & $<0.001$ & $1.82$ \\
      \bottomrule
    \end{tabular}
    \begin{tablenotes}
      \footnotesize
      \item Scores are mean $\pm$ standard deviation on a 1--5 Likert scale.
      \item All dimensions were tested with the two-sided Wilcoxon signed-rank test (Shapiro--Wilk $p<0.05$).
      \item $p_{\mathrm{adj}}$: Benjamini--Hochberg FDR-corrected $p$-values.
    \end{tablenotes}
  \end{threeparttable}
\end{table*}

%% file: paper_tables/rq3_cognitive_load.tex
\begin{table*}[pos=tbp]
  \centering
  \footnotesize
  \caption{RQ3.1: Cognitive load comparison between text-only and multimodal treatment plans (LLM-as-a-Judge, $N=30$).}
  \label{tab:rq3-cognitive}
  \begin{threeparttable}
    \begin{tabular}{lcccccc}
      \toprule
      Dimension & Plan A (text) & Plan B (multimodal) & $p$ & $p_{\mathrm{adj}}$ & $r$ & $\Delta$ \\
      \midrule
      Language comprehensibility & $2.10 \pm 0.40$ & $2.60 \pm 0.62$ & $<0.001$ & $<0.001$ & $0.88$ & $+0.50$ \\
      Preparation clarity & $3.40 \pm 0.67$ & $3.67 \pm 0.76$ & $0.046$ & $0.053$ & $0.57$ & $+0.27$ \\
      Acupoint locatability & $1.63 \pm 0.56$ & $3.50 \pm 0.57$ & $<0.001$ & $<0.001$ & $1.00$ & $+1.87$ \\
      Action confidence & $1.97 \pm 0.32$ & $2.67 \pm 0.61$ & $<0.001$ & $<0.001$ & $0.91$ & $+0.70$ \\
      \midrule
      Overall cognitive load\tnote{a} & $1.43 \pm 0.50$ & $2.03 \pm 0.72$ & $<0.001$ & $<0.001$ & $0.90$ & $+0.60$ \\
      \bottomrule
    \end{tabular}
    \begin{tablenotes}
      \footnotesize
      \item Scores are mean $\pm$ standard deviation on a 1--5 Likert scale.
      \item All dimensions were tested with the two-sided Wilcoxon signed-rank test; $r$ is the rank-biserial effect size.
      \item $p_{\mathrm{adj}}$: Benjamini--Hochberg FDR-corrected $p$-values. $\Delta$ = Plan~B mean $-$ Plan~A mean; positive values favor Plan~B.
      \item[a] Reverse scored: higher = lower cognitive load (better).
    \end{tablenotes}
  \end{threeparttable}
\end{table*}

%% file: paper_tables/rq3_references.tex
\begin{table*}[pos=tbp]
  \centering
  \footnotesize
  \caption{RQ3.2: Impact of structured references on perceived evidence quality (LLM-as-a-Judge, $N=25$).}
  \label{tab:rq3-references}
  \begin{threeparttable}
    \begin{tabular}{lccccc}
      \toprule
      Dimension & With references & Without references & $p$ & $p_{\mathrm{adj}}$ & Cohen's $d$ \\
      \midrule
      Source credibility & $4.11 \pm 1.18$ & $2.60 \pm 0.53$ & $<0.001$ & $<0.001$ & $1.52$ \\
      Evidence verifiability & $3.86 \pm 1.30$ & $1.30 \pm 0.00$ & $<0.001$ & $<0.001$ & $1.98$ \\
      Information confidence & $4.48 \pm 1.12$ & $3.73 \pm 0.84$ & $<0.001$ & $<0.001$ & $1.43$ \\
      Comprehension support & $4.61 \pm 1.10$ & $4.66 \pm 1.03$ & $0.317$ & $0.317$ & $-0.20$ \\
      Overall evidence quality & $4.00 \pm 1.30$ & $2.44 \pm 0.43$ & $<0.001$ & $<0.001$ & $1.55$ \\
      \midrule
      Overall mean & $4.21$ & $2.95$ & --- & --- & $\Delta=+1.27$ \\
      \bottomrule
    \end{tabular}
    \begin{tablenotes}
      \footnotesize
      \item Scores are mean $\pm$ standard deviation on a 1--5 Likert scale.
      \item All dimensions were tested with the two-sided Wilcoxon signed-rank test.
      \item $p_{\mathrm{adj}}$: Benjamini--Hochberg FDR-corrected $p$-values. $d$: paired-sample Cohen's $d$.
    \end{tablenotes}
  \end{threeparttable}
\end{table*}

%% file: sections/discussion.tex

This section discusses design implications derived from the system's design and evaluation (Section~\ref{sec:design-implications}), clinical integration considerations (Section~\ref{sec:clinical-integration}), and limitations (Section~\ref{sec:limitations}).

\subsection{Design Implications and Responses to Research Questions}\label{sec:design-implications}

\paragraph{I1: Soft Scaffold, Hard Boundary --- KG Constraints Meet Multi-Round Interaction (RQ1, C2)}
Our system uses the KG simultaneously as a ``soft scaffold'' guiding interaction and a ``hard boundary'' constraining outputs. At the interaction level, the information gain-driven follow-up strategy selects the most discriminative symptoms within the candidate space, turning diagnosis into a short guided dialogue rather than a single-shot query~\cite{11,12}; this is a design property illustrated by the case walkthrough in Supplementary Information Section~S3, not a quantified result. At the constraint level, the LLM handles flexible symptom extraction on the input side, while the KG constrains the output space to validated syndrome-symptom associations on the output side, reducing the risk of hallucinated diagnostic conclusions~\cite{25,64,44,45}. In evaluation, the KG-grounded system achieved higher Top-1 accuracy than both LLM baselines; because that comparison is system-level rather than an ablation, the margin cannot be attributed to the graph constraint alone. This complementary architecture addresses the respective limitations of KG-only rigidity and LLM-only unreliability.

\paragraph{I2: Diagnosis as Explanation --- Progressive Visualization Builds Trust (RQ2)}
The system's real-time progressive KG updating transforms the diagnostic process itself into an explanatory tool: users observe how candidate syndromes narrow across interaction rounds, understanding the basis behind each follow-up question. The diagnosis-treatment dual-chain visualization further enables users to trace reasoning paths from symptoms to syndrome patterns and onward to treatments, making conclusions traceable and verifiable. This design implies that health informatics systems should treat the diagnostic process, not merely the final decision, as the core object of visualization~\cite{24,35}. A transparent reasoning process is an important factor for building trust~\cite{40}, although transparency alone is not sufficient to address all user risks and must be complemented by contextualized usage guidelines~\cite{41}. Evaluation confirmed that this approach produced a large effect improvement across all five trust dimensions.

\paragraph{I3: Heterogeneous Modality Division --- Multimodal Presentation Lowers Cognitive Barriers (RQ3)}
Different modalities serve different cognitive functions: text conveys logical reasoning, images convey intuitive understanding, and 3D models convey spatial localization~\cite{67,Baxter2025}. This division enables differentiated information access: patients grasp treatment essentials through intuitive images, physicians consult detailed formula compositions and KG reasoning paths, and students use interactive 3D models for acupoint localization~\cite{30,69}. Evaluation confirmed that the multimodal plan significantly reduced cognitive load compared to text-only presentation, with acupoint locatability showing the most pronounced improvement.

\subsection{Clinical Integration Considerations}\label{sec:clinical-integration}

The system should function as a decision-support tool embedded within physicians' existing workflows, with final decisions made by licensed physicians. Deployment requires professional training, informed consent regarding AI involvement, and data privacy compliance~\cite{7,8,14}.

\subsection{Limitations}\label{sec:limitations}

We identify the following limitations of the current system:

\begin{enumerate}
  \item \textbf{Limited KG coverage}: The knowledge graph covers 241 syndrome patterns, but TCM literature documents over 1,000 types; rare syndromes and regional differentiation systems (e.g., Lingnan medicine) remain excluded.
  \item \textbf{Lack of large-scale user study}: Evaluation relied on the LLM-as-a-Judge framework with limited sample sizes ($N=25$--$30$), which may exhibit systematic biases. A controlled user study with practitioners remains necessary for validating clinical utility.
  \item \textbf{Questioning policy not ablated}: The follow-up strategy driven by information gain is demonstrated in a case walkthrough only. Convergence efficiency (rounds to convergence, candidate-space reduction per round) was not measured, so its contribution cannot be separated from that of the knowledge graph constraint evaluated under RQ1.
  \item \textbf{LLM content reliability}: Although KG constraints reduced non-standardized outputs by 32\%, treatment plans may still contain inaccurate recommendations. The system currently lacks a human-AI collaborative review mechanism for physician oversight.
  \item \textbf{No real clinical validation}: Evaluation used retrospective CEMRs data without prospective validation in real clinical settings. A prospective clinical trial is necessary to assess real-world performance.
\end{enumerate}

%% file: sections/conclusion.tex

This paper presented a knowledge graph-driven TCM visual diagnostic system that addresses three interrelated challenges in AI-assisted syndrome differentiation: opaque reasoning, passive interaction, and simplistic treatment presentation. By combining progressive KG visualization with LLM-based symptom extraction, the system makes diagnostic reasoning transparent and interactive while constraining outputs to validated medical knowledge. Evaluation demonstrated substantial trust improvements (Cohen's $d=1.82$) and reduced cognitive load through multimodal treatment plans. KG constraints also reduced non-standardized LLM outputs by 32\%. These results suggest that structured knowledge constraints can complement generative AI capabilities in clinical decision support. Current limitations include reliance on automated evaluation with limited sample sizes and incomplete KG coverage. Future work will prioritize prospective clinical validation with practitioners and patients, expanded syndrome coverage, and EHR integration to support real-world deployment.

%% file: sections/declarations.tex
\section*{Ethics statement}
This study focuses on system design and technical implementation. It does not involve human experimentation or patient data collection, and no ethics review approval was required. Expert consultations during the system evaluation phase were conducted for the purpose of technical functionality verification and did not involve clinical intervention.

\section*{Funding}
This work was supported in part by the National Natural Science Foundation of China (Grant No. 62472121), the Key Technology Research and Development Program of Shandong Province (Grant No. 2025CXPT077), and the Special Funding Program of Shandong Taishan Scholars Project.

\section*{Declaration of competing interest}
All authors declare no conflicts of interest.

\section*{Data availability}
Due to privacy restrictions, the complete CEMRs dataset cannot be publicly released. A de-identified sample dataset and complete prompt templates will be provided as supplementary material for independent validation.

\section*{Declaration of generative AI and AI-assisted technologies in the manuscript preparation process}
During the preparation of this work, the authors used OpenAI Codex for LaTeX template migration and formatting support. After using this tool, the authors reviewed and edited the content as needed and take full responsibility for the content of the article.